A Visual Diagnostics Framework for District Heating Data: Enhancing Data Quality for AI-Driven Heat Consumption Prediction

Christensen, Kristoffer; Jørgensen, Bo Nørregaard; Ma, Zheng Grace

Go to publication entry in University of Southern Denmark's Research Portal

# A Visual Diagnostics Framework for District Heating Data: Enhancing Data Quality for AI-Driven Heat Consumption Prediction

Kristoffer Christensen[1]*[0000-0003-2417-338X], Bo Nørregaard Jørgensen[1][0000-0001-5678-6602] and Zheng Grace Ma[1][0000-0002-9134-1032]

[1] SDU Center for Energy Informatics, Maersk Mc-Kinney Moeller Institute, The Faculty of Engineering, University of Southern Denmark, Odense, Denmark

`kric@mmmi.sdu.dk, bnj@mmmi.sdu.dk and zma@mmmi.sdu.dk`

**Abstract.** High-quality data is a prerequisite for training reliable Artificial Intelligence (AI) models in the energy domain. In district heating networks, sensor and metering data often suffer from noise, missing values, and temporal inconsistencies, which can significantly degrade model performance. This paper presents a systematic approach for evaluating and improving data quality using visual diagnostics, implemented through an interactive web-based dashboard. The dashboard employs Python-based visualization techniques, including time series plots, heatmaps, box plots, histograms, correlation matrices, and anomaly-sensitive KPIs such as skewness and anomaly detection based on the modified z-scores. These tools allow human experts to inspect and interpret data anomalies, enabling a human-in-the-loop strategy for data quality assessment. The methodology is demonstrated on a real-world dataset from a Danish district heating provider, covering over four years of hourly data from nearly 7000 meters. The findings show how visual analytics can uncover systemic data issues and, in the future, guide data cleaning strategies that enhance the accuracy, stability, and generalizability of Long Short-Term Memory and Gated Recurrent Unit models for heat demand forecasting. The study contributes to a scalable, generalizable framework for visual data inspection and underlines the critical role of data quality in AI-driven energy management systems.

## 1    Introduction

District Heating (DH) networks are a cornerstone of Denmark's energy system, supplying heat to a majority of households. Approximately 66% of Denmark's heating demand is met by district heating networks [1]. This extensive coverage generates a wealth of operational data from heat meters at buildings and substations. These data are increasingly used to train Artificial Intelligence (AI) models for forecasting heat consumption, enabling better load balancing and energy efficiency. However, real-world district heating data often suffers from quality issues such as missing readings, anomalous spikes or drops, and inconsistent records. If left unaddressed, such issues can significantly impair the performance of data-driven models. Before deploying advanced AI models like Long Short-Term Memory (LSTM) and Gated Recurrent Unit (GRU) networks to predict building-level heat consumption, it is crucial to ensure the training data is clean and reliable. The focus of this paper is on visual diagnostics as a strategy for improving data quality in Danish district heating systems. This work demonstrates how data visualization techniques can help engineers and analysts identify and correct data issues (e.g., missing values and anomalies) prior to model training. By applying visual analytics to a district heating dataset, data integrity is enhanced, resulting in more accurate and trustworthy AI-driven predictions of heat consumption. Furthermore, a case study is presented highlighting common data problems and their visual signatures and discussing the implications of data quality on training LSTM/GRU models for heat demand forecasting.

This study is guided by the overarching question: *How can visual analytics be used to systematically evaluate and improve the quality of district heating data prior to AI model training*? To address this, the following sub-questions are investigated:

1. What types of data quality issues are most prevalent in real-world district heating metering data, and how can they be identified visually?
2. How can a structured set of key performance indicators (KPIs) and interactive visual tools support the diagnosis of such issues?
3. In what ways do visual diagnostics inform data cleaning decisions that enhance the performance and stability of AI models, particularly recurrent neural networks for heat consumption prediction?

To answer these questions, this paper makes the following contributions:

1. It proposes a dashboard-based visual analytics framework for data quality assessment in district heating systems, incorporating interactive time series plots, heatmaps, statistical distributions, and correlation analyses.
2. It introduces a set of domain-specific KPIs, including statistics on anomalies, skewness, and null values, visualized using gauge components to guide interpretation.
3. It applies the framework to a real-world dataset from a Danish district heating provider, uncovering common data quality issues such as seasonal dropout patterns, extreme outliers, and negative consumption values.
4. It discusses the implications of these diagnostics for training LSTM and GRU models, demonstrating how visual inspection supports model reliability and data trustworthiness.
5. It positions visual diagnostics as a scalable, human-in-the-loop strategy for improving data quality in smart energy systems, with the potential for generalization to other time-series domains such as electricity, water, and industrial sensor networks.

The paper is organized as follows. Section 2 reviews related work; Section 3 presents visual diagnostics and indicators; Section 4 describes the data pipeline and ar-chitecture; Section 5 reports a Danish case study; Section 6 discusses implications for LSTM- and GRU-based models; Section 7 reflects on human-in-the-loop diagnostics; and Section 8 concludes with future work.

## 2 Background and Related Work

**District Heating in Denmark**: Denmark's DH infrastructure is widely regarded as a mature and efficient system. The extensive rollout of smart heat meters in recent years has produced large datasets of hourly consumption from thousands of buildings [2]. For instance, Schaffer et al. published a dataset of three years of hourly heat consumption from 3021 Danish residential buildings [3]. These datasets provide a basis for advanced analytics but also underscore the importance of data cleaning. In the Aalborg city dataset, careful preprocessing was needed to remove errors and fill gaps, resulting in a final dataset with virtually no missing values [2]. Such efforts illustrate the growing recognition that data quality must be managed proactively in the energy domain.

**Heat Consumption Forecasting**: Accurate forecasting of building heat demand is vital for optimizing DH system operations and integrating renewable sources. Traditional approaches included statistical models and engineering simulations, but recent years have seen a surge in machine learning methods [4]. Deep learning models, especially recurrent neural networks like LSTMs and GRUs, have shown promise in capturing complex temporal patterns of heat usage. Golmohamadi (2022) demonstrated that data-driven models leveraging weather inputs can successfully predict building heat consumption, with LSTM networks achieving high short-term accuracy [4]. Huang et al. (2023) introduced an explainable Graph Recurrent Network that outperformed eleven baseline models on a Danish heating dataset [5], highlighting the state-of-the-art in DH load forecasting with AI. Meanwhile, Kristensen et al. (2020) explored a complementary approach using hierarchical archetype modeling for long-term (annual) heat load forecasting of urban areas [4]. These works collectively show that AI techniques, especially LSTM/GRU-based models, are increasingly central to DH consumption prediction.

**Data Quality Challenges**: Despite these advances, issues of data quality remain a limiting factor. Many forecasting studies assume well-prepared data, yet operational DH datasets are prone to errors. Common problems include sensor outages leading to missing intervals, meter malfunctions causing negative or zero readings, and occasional extreme spikes unrelated to actual consumption (e.g., due to telemetry errors). Zangrando et al. (2022) investigated anomaly detection in energy consumption series and found that even sophisticated deep learning methods (e.g., LSTM autoencoders) can struggle with generalizing to unseen anomalies, sometimes being outperformed by simpler methods for outlier detection [6]. This finding underscores the need for robust preprocessing: if anomalies are not handled prior to training, complex models may not inherently "learn around" them and could be misled by spurious patterns. In practice, energy utilities often apply rule-based filters or basic statistical checks, but these may not catch all issues or might remove legitimate but unusual behaviors. A more nuanced approach is required to ensure data quality without discarding useful information.

**Visual Analytics for Data Quality**: Visual analytics has emerged as a powerful approach to assess and improve data quality in machine learning pipelines. Liu et al. (2018) advocate "steering" data quality through interactive visualization, arguing that human insight is crucial to navigate the complexity of real-world data issues [7]. By visualizing data distributions, time series, and anomalies, analysts can detect patterns that automated algorithms might miss (or misclassify). In the context of building energy data, visual inspection allows domain experts to differentiate between true outliers and explainable variations (for example, a holiday period with atypical consumption versus a sensor fault). This paper builds on that premise, applying visual diagnostic techniques specifically to DH consumption data. This paper extends prior work by demonstrating concrete visualization methods implemented in an interactive web-based dashboard. The dashboard is developed in Python using Dash and Plotly as the main libraries for interactive visualization.

Recent advances in visual analytics highlight the critical role of domain expertise in interpreting complex patterns in real-world sensor data. Unlike fully automated data cleaning techniques, visual analytics enables human analysts to apply contextual knowledge (such as operational schedules, expected consumption trends, or known system behaviors) to distinguish between true anomalies and explainable variations. This human-in-the-loop approach is especially important in time-series anomaly detection, where subtle contextual cues or systemic data dropouts may elude algorithmic filters. Research in visual analytics communities such as IEEE VIS and EuroVis has demonstrated the effectiveness of interactive techniques for anomaly exploration in temporal data. For instance, methods like horizon graphs, multi-resolution time series brushing, and context-aware heatmaps have been shown to support the discovery of both localized and structural anomalies. These works underscore that visual interfaces not only facilitate error detection but also foster data trust and transparency, key prerequisites for deploying AI models in high-stakes domains such as energy systems. While this paper emphasizes human-in-the-loop visual analytics, recent perspectives on agentic AI Tirulo et al. (2026) argue that autonomous agents may increasingly take on such tasks, reducing reliance on human oversight [8]. Yet, in the context of time-series sensor data, their application to data quality diagnostics is still largely prospective rather than empirically established, making human expertise currently indispensable.

## 3 Visualization Techniques for Data Quality Assessment

A range of Python-based visualization techniques was applied to diagnose data quality issues in the heating consumption data. The following methods proved especially useful:

**Gauge indicators**: For a dashboard design with an overall overview, Key Performance Indicators (KPIs) based on data quality metrics are essential for quick assessment of the data quality. The gauge visualizations are driven by statistical KPIs de-signed to help analysts assess the robustness and quality of data prior to its use in AI-based forecasting methods, utilizing color codes: Green, Yellow, and Red to visualize the quality or robustness for each KPI. The purpose of the colors and the different KPIs are explained as follows: Green: Normal/healthy data, no immediate concern; Yellow: The user should monitor, investigate; Red: Detailed analysis is needed to confirm quality or anomalies

**Mean and Median**: A gauge indicator is used to display the mean value on a scale spanning from the minimum to the maximum of the dataset. The gauge employs a symmetrical color gradient centered around the middle range, designed to visually convey data distribution characteristics. Specifically, the central zone (30–70% of the scale) is marked in green, indicating a balanced distribution. This is flanked by yellow bands (each covering 7.5%) and red bands (each covering the outermost 22.5%), extending toward both ends of the scale. The color scheme is intended to highlight potential data quality concerns: for instance, a needle positioned within the red zone may suggest the presence of significant outliers or skewed data. As such, the mean alone may not provide a complete picture and is therefore complemented by a box plot to better assess distribution symmetry and the presence of anomalies.

**Median Absolute Deviation**: The MAD is a robust measure of variability and is calculated as shown in equation 1 [9]:

$$MAD = median_i\{|x_i - \tilde{x}|\}$$ (1)

Where $x_i$ is each individual data point and $\tilde{x}$ is the median of the dataset. It shows how much the values in a dataset typically differ from the median. Unlike standard deviation, MAD is less affected by outliers and skewed data, making it useful for understanding the spread of real-world, noisy datasets, as is often the case for sensor data [10]. A low MAD means most values are close to the median; a high MAD indicates greater spread. The MAD is further used in the calculation of the modified z-score.

**Anomaly Count Based on Absolute Modified Z-score**: This metric represents the anomaly count based on the absolute modified z-score value. The modified z-score quantifies how many robust standard deviations a value deviates from the robust center, typically the median, using robust measures of spread such as the MAD instead of the mean and standard deviation. The calculation of the modified z-score is detailed in Equation 2 [9]:

$$M_i = \frac{0.6745(x_i - \tilde{x})}{MAD}$$ (2)

Where $M_i$ is the modified z-score of data point $x_i$. The robustness of using the modified z-score is particularly valuable in sensor data, which often exhibits spikes, noise, or outliers due to measurement errors or environmental disturbances. Unlike the conventional z-score, the modified z-score minimizes the influence of such irregularities, providing a more reliable characterization of both typical and extreme data behavior. Accordingly, this makes it a more suitable metric for anomaly detection in noisy environments. A value is flagged as an anomaly when the absolute modified z-score exceeds a threshold of 3.5, reflecting widely adopted thresholds in anomaly detection literature [9, 11]. The associated color scale follows the acceptable threshold for anomalies in a dataset of <5% [9, 12]. Hence, an anomaly counts of less than 5% is green, 5-10% is yellow, and all above is red.

**Number and percentage of null values**: A metric showing if missing values are a major issue for the investigated dataset. While no universally accepted standard exists for acceptable missing data, many experts suggest that tolerating less than 5% missing values is generally manageable in predictive modeling [13]. At the other extreme, variables with more than 50% missing data are often excluded, as imputing them may introduce substantial noise and degrade data quality [14]. Accordingly, the traffic-light approach is used for data quality: <5% missing as acceptable, 5–50% requiring careful handling, and >50% flagged as severely incomplete.

**Robust Skewness (Medcouple)**: The medcouple (MC) is a robust measure of skewness that quantifies distributional asymmetry while reducing sensitivity to outliers [15]. It is defined as shown in Equation 3 and 4:

$$MC = \underset{x_i \leq m_n \leq x_j}{med} h(x_i, x_j) \tag{3}$$

$$h(x_i, x_j) = \frac{(x_j - m) - (m - x_i)}{x_j - x_i} \tag{4}$$

where $m$ is the median of the dataset, and $x_i$, $x_j$ are elements from the lower and upper halves, respectively [15]. In practice, values lie within –1 to +1, with positive MC indicating right-skewness and negative MC left-skewness. For visualization, the heuristic thresholds are defined: green for approximately symmetric data ($-0.2 \leq MC \leq 0.2$), yellow for moderate skew ($0.2 < |MC| \leq 0.5$), and red for strong skew ($|MC| > 0.5$). These thresholds are not universally accepted statistical cutoffs but are empirically chosen to provide intuitive traffic-light feedback on sensor data quality and asymmetry.

**Box-plot**: For visual indication of outliers, box plots are great as they contain maximum, minimum, mean, median, first- and third quartiles in one figure.

**Time-Series Line Plots**: The KPI overview might result in the need for a deeper understanding of the data, for which the plotting of the time series data is a great tool. The plot should allow for interactions, such as hovering data points to access content, period selection, and zooming in to understand the anomalies. Continuous line plots make missing data evident as breaks in the line, and they highlight sudden spikes or drops.

**Data-grid**: With data for many thousands of meters, a data-grid offers customizable filtering, sorting, grouping, etc. in a user-friendly and visual way, not requiring any coding. This enables easy filtering of meters based on specific data statistics, such as a percentage of null values above 90%, helping to identify faulty meters. These meters can then be exported as a list for further review or exclusion.

**Heatmap of Time vs. Time**: A two-dimensional heatmap, with one axis representing the hour of the day and the other axis representing the date (or day of year), provides a compact view of data completeness and periodicity. An example of such a figure could be hours of the day (y-axis) versus time progressing over years (x-axis) for a DH dataset, with colors indicating how many meters have missing data at a giv-en hour. Such a visualization can uncover systemic data dropouts – for instance, vertical bands in the heatmap correspond to specific dates where many meters lost data simultaneously [3]. Heatmaps with colors indicating the value of a certain reading, e.g., heat consumption, are currently not considered for identifying data quality, since extreme outliers dominate the color scaling. Such a heatmap is relevant after removing extreme outliers.

**Histogram**: Histograms are used to examine the distribution of data values across different ranges or bins. By visualizing how frequently values occur within these intervals, histograms help detect patterns such as skewness or multimodality in the data. This makes it easier to identify outliers, especially extreme high values that may indicate anomalies, measurement errors, or faulty sensors. Additionally, histograms can reveal physically impossible or unexpected values, providing a quick diagnostic tool to assess data quality and integrity.

**Dynamic scatter plot and correlation matrix**: Plotting numerical data columns against each other in a scatter plot can help identify patterns such as clustering, trends, or potential anomalies. The dashboard allows users to dynamically select which columns to compare via dropdown menus, providing interactive and flexible exploration.

A correlation matrix offers a fast and intuitive overview of relationships between all numerical columns. It is especially useful for detecting linear dependencies or unexpected deviations, which may signal faulty or anomalous sensor readings.

To enhance this analysis, the dashboard supports both Pearson and Spearman correlation methods, selectable through a dropdown.

- Pearson correlation measures the strength of linear relationships between variables and is most appropriate when data is normally distributed and free of significant outliers.
- Spearman correlation, on the other hand, evaluates monotonic relationships using rank order and is more robust to non-linear patterns and outliers.

This flexibility allows users to choose the correlation method best suited to the data characteristics, improving the reliability of anomaly detection and data quality assessment

**Interactive Visualization**: An interactive time series plot allows panning/zooming to investigate specific dates, and tooltips can display exact values to quantify anomalies. Such interactivity is especially useful when dealing with overlaying plots or very long time series, where filtering by trace or date range can isolate issues for closer examination.

To summarize, Table 1 maps each visualization diagnostics technique to the related issues, metrics, and suggested actions. The suggested actions indicate potential next steps; in the current version, the tool does not perform data cleaning but only flags anomalies. Detailed cleaning strategies, including removal or imputation, are planned for future work.

**Table 1.** Mapping of Visual Diagnostics to Data Quality Issues, KPIs, and Actions.

| Visualization | Issue Detected | Metric / KPI | Suggested Action |
|---|---|---|---|
| **Time-Series Line Plot** | Sudden spikes, drops, and missing values | Visual inspection (breaks or discontinuities) | Investigate anomalies, impute, or remove extreme values |
| **Box Plot** | Outliers, skewed distributions | Mean vs. Median, Quartiles | Confirm and flag outliers, assess distributional bias |
| **Histogram** | Extreme values, non-physical readings | Frequency distribution | Clip or correct invalid values (e.g., negative energy) |
| **Heatmap (Time vs. Time)** | Systemic missing patterns, temporal dropout | Percentage of null values | Identify periods with high loss; flag or impute |
| **KPI Gauges (e.g., Anomaly, Nulls)** | Data spread, presence of missing or extreme values | Anomaly, Skewness, Null % | Determine data health; prioritize columns/segments for review |
| **Data Grid (Tabular View)** | Unexpected meter behavior | All mentioned, e.g., Null % | Filter and isolate problematic meters for further inspection |
| **Scatter Plot (Dynamic)** | Abnormal inter-feature relationships | Visual deviation from expected correlations | Identify and exclude implausible points or faulty sensors |
| **Correlation Matrix** | Lack of correlation, variable redundancy, or error | Pearson/Spearman correlation | Validate expected relationships; detect systemic anomalies |

## 4 Data Pipeline and Visualization Architecture

The data pipeline architecture used for the dashboard visualization is illustrated in Fig. 1. The data is collected from district heating meters operated by the district heating provider and ingested into their database. Currently, for third-party analysts, the data is extracted and shared as csv files. Due to the size of data, also for future applications, DuckDB has been selected as the database SQL database management system. DuckDB is an embedded, in-process SQL database optimized for analytical workloads (OLAP) [16]. It combines the simplicity of SQLite with the performance of modern columnar databases. Key advantages include: No server required: Fully embedded and portable with zero setup or dependencies; High performance: Uses a vectorized columnar engine for fast analytical queries on large datasets; Python/R integration: Can query Pandas data directly, avoiding copies or ETL; Rich SQL support: ACID compliance, window functions, and secondary indexes; Extensible: Supports plugins for Parquet, JSON, S3, and more; Cross-platform: Runs on everything from servers to browsers (via DuckDB-Wasm); Open source: MIT-licensed, community-driven, and thoroughly tested. DuckDB is ideal for local analytics, dashboards, data science workflows, and anywhere fast serverless OLAP is needed [16].

For further processing of the resulting data frames extracted using DuckDB, Pandas, and Numpy are used in the data processing. Data frames extracted from larger analytic processing are saved locally as parquet files to ensure low memory usage and fast processing in the following executions.

Dash is used as the tool for developing a customized interactive web-based dashboard [17]. Dash is the most suitable choice for the dashboard requirements: it is native to Python, seamlessly integrates with the scientific computing ecosystem (such as Pandas, Plotly, and NumPy), enables scalable deployment via web-based rendering technologies (HTML/CSS/JavaScript), and allows for extensive customization through its callback-based logic and modular components [18]. Dash apps mainly consist of two core components: 1) Layout, defines the structure and appearance of the app, built using Python objects that map to HTML and CSS elements via the dash.html and dash.dcc libraries; 2) Callback functions, add interactivity by connecting inputs (e.g., dropdowns, sliders) to outputs (e.g., graphs, text), allowing the app to respond to user actions in real time. Both components are represented in the architecture design.

All visualizations were implemented in Python, using Plotly for interactive plots. The ease of plotting with these libraries enabled an iterative process: identify an anomaly visually, correct it, and re-check the data after modifications. Future extensions of the program should include direct modifications to the data to evaluate the impact directly and export the finalized data.

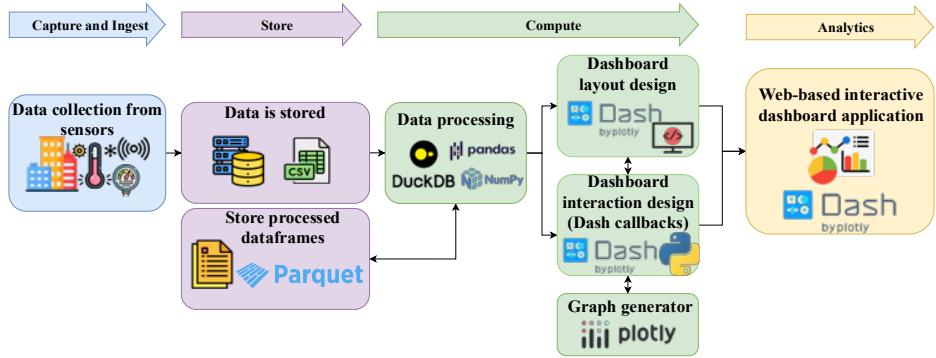

**Fig. 1.** Data pipeline architecture for data visualization.

## 5 Case Study: Visual Data Cleaning of Heat Consumption Data

A case study has been selected for a district heating area within the area operated by the Danish district heating provider, Trefor. The data is hourly readings from January 1st, 2020, to September 4th, 2024, with 6923 unique metering IDs. Besides the timestamp, the data contains measurements for energy (MWh), forward temperature (°C), return temperature (°C), flow (L/h), and energy computed (MWh).

Running the data through the pipeline and running the dashboard application, metrics for each measurement are found and visualized as seen in Fig. 2.

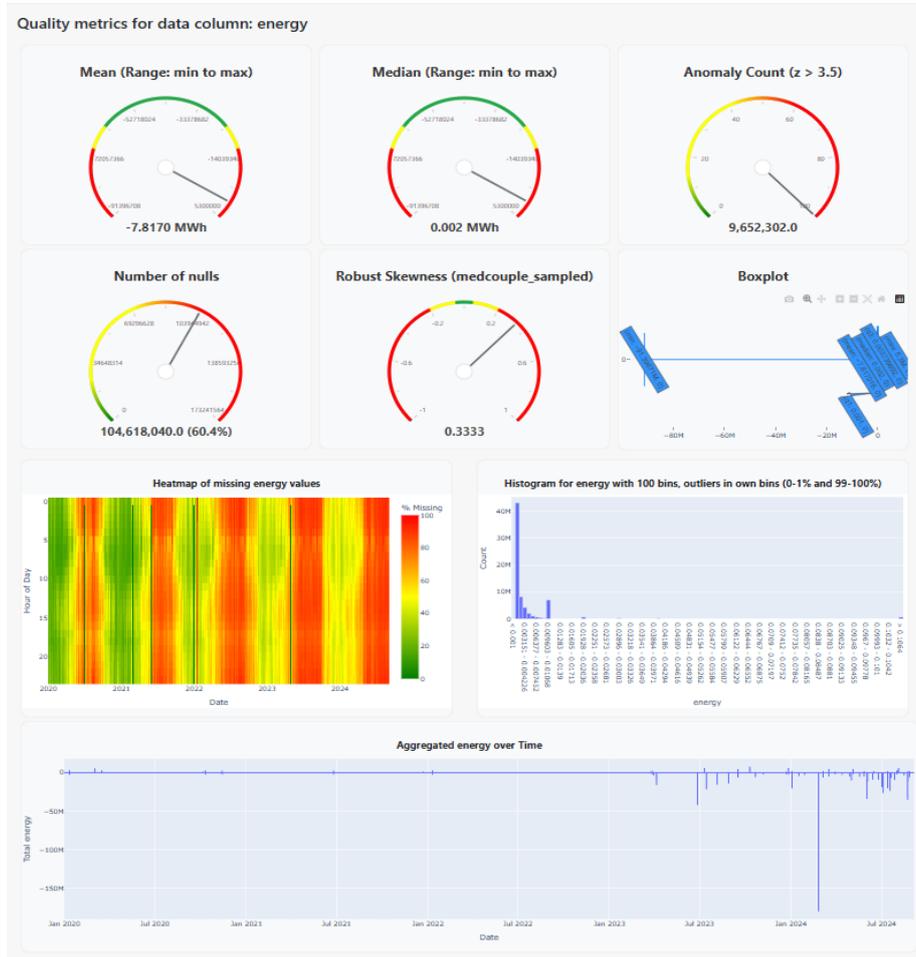

**Fig. 2.** Dashboard overview screenshot for measurements using "energy" as an example. Includes gauges for mean, median, anomaly count, null count, robust skewness, boxplot, heatmap of missing values, histogram, and line plot.

The figure only shows energy measurements as an example. All figures, besides the gauges, are interactive, enabling zoom, hovering information, changing axes ranges, and the selection of specific periods. Without any closer look, it is evident from the figure that this data is heavily impacted by extreme outliers and many missing values. 60% of the data is missing, and from the heatmap, it seems that there is a correlation between season and the number of missing values. The high percentage of missing values is repeated every summer season and might indicate that meters or systems are turned off. If this can be confirmed by an operations expert, a strategy for how to detect such occurrences and convert values to zero should be considered. All the gauges besides the number of nulls indicate that extreme outliers are present, and also confirmed by the box plot, that are dominated by the outliers, so the full box plot is not visible. That is further confirmed by the histogram and line plot. Furthermore, the line plot shows a lot of negative values, which, by inclusion of domain rules, can easily be removed. Assuming the meters only represent consumers, the energy can never be negative.

In another tab in the dashboard, for data analysis, a field for data correlation is designed as shown in Fig. 3. From the correlation matrix, flow has the strongest correlation with energy and energy computed. The correlation is visualized in the scatterplot, showing a linear correlation. For fast processing, rendering, and responsiveness, a maximum of 100,000 datapoints is sampled in the figure.

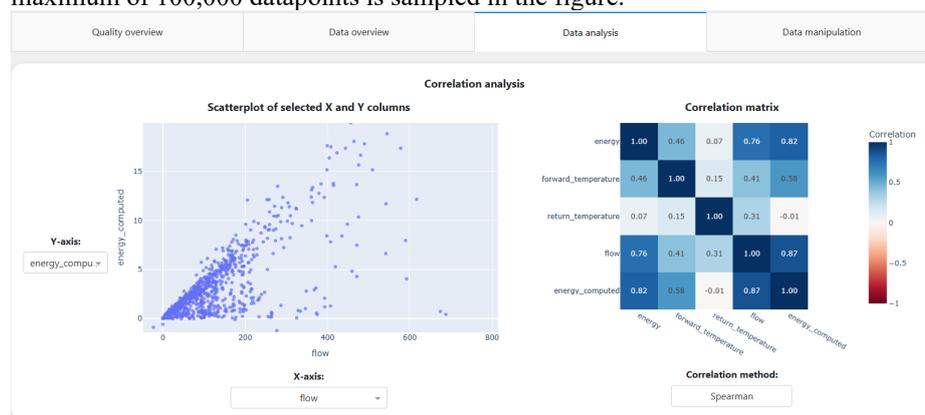

**Fig. 3.** Correlation section with dynamic scatterplot and correlation matrix for all numerical measurements in the dataset.

Fig. 4 shows the heatmap of missing metering IDs. From the figure, it is evident that more meters are added to the dataset over time and that there are several days with zero meters, with a general pattern of being the last day of the month until 2023 when it disappears. This explains the same pattern found in the missing energy values in Fig. 2, showing no missing values, due to no meters present at that time. A total of 943 timestamps are missing in the dataset period. This again requires a dialog with the meter operators to understand the reason.

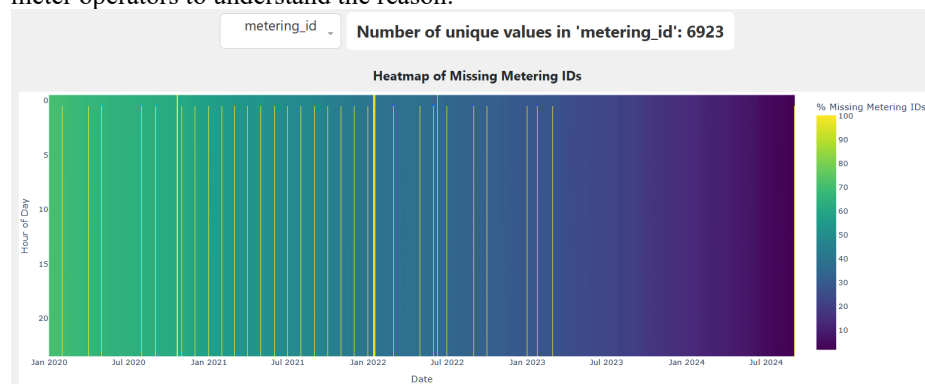

**Fig. 4.** Heatmap of missing metering IDs within the day over time.

# 6 Implications for AI Model Training

High-quality data is the foundation for effective machine learning models. The LSTM and GRU architectures, in particular, are powerful at capturing temporal dependencies but can be sensitive to noise and irregularities in training data. The visual data cleaning process has several important implications for training these models to predict building-level heat consumption:

**Improved Model Accuracy**: By removing extreme outliers and filling gaps, the training data more accurately represents the true heat consumption patterns. As a result, the LSTM/GRU can learn the relationship between inputs and future consumption without being misled by spurious events. If an extreme spike remained in the data, the model might try to fit it, allocating capacity to an event that will not generalize. Empirically, cleaning often leads to lower error on validation data; although results vary, reductions are expected in Mean Absolute Error (MAE) and Root Mean Square Error (RMSE) after cleaning since the model no longer chases noise. This aligns with the observations of Zangrando et al. that removing or properly labeling anomalies can yield simpler models that outperform more complex ones on "dirty" data [6]. It should be noted that this section is conceptual in nature; the claims are grounded in prior research and domain logic rather than derived from a direct experimental comparison. A future study could empirically validate these implications by training LSTM/GRU models on both cleaned and uncleaned versions of the same district heating dataset and comparing forecasting metrics such as RMSE and MAE.

**Training Convergence and Stability**: LSTMs and GRUs are trained with gradient-based optimization. Large anomalies can cause large gradient updates, potentially destabilizing training (e.g., causing exploding gradients or oscillations in the loss curve). By capping or eliminating extreme values, the risk of such training pathologies is reduced.

**Reduced Overfitting to Errors**: If not removed, anomalies can inadvertently become easy targets for the model to overfit. For instance, a lone high spike might be memorized by the network, drastically reducing training errors but obviously not helping real predictive power. By eliminating these artifacts, it is encouraged for the model to generalize on the true underlying patterns (like diurnal and temperature-driven patterns) rather than overfitting to idiosyncrasies.

**Handling of Missing Data**: Sequence models like LSTM and GRU cannot inherently handle missing timesteps; typically, missing data must be imputed or masked. By using visual diagnostics to decide on an imputation strategy (or to decide that certain segments should be entirely omitted), it is ensured the model is never fed undefined inputs.

**Feature Engineering Confidence**: With anomalies handled, features for the model can be more confidently developed. For example, creating lag features (previous day consumption, etc.), knowing that those values are real or reasonable estimates, not error artifacts. Similarly, scaling/normalizing the inputs (a common step for LSTM/GRU) is safer when outliers are removed – otherwise, an outlier could skew the scaling of all data (e.g., if using min-max normalization, a single spike would flatten the scale of all other inputs).

**Model Architecture and Hyperparameters**: Confidence in data cleanliness can influence modeling choices. For example, a smaller network or less regularization may be selected when the data is trusted because there is less concern about the model's learning noise. On the other hand, if minor anomalies remain or the data is naturally noisy due to variations in occupant behavior, robust training strategies such as using Huber loss instead of MSE to reduce the impact of outliers might be used.

Importantly, these implications extend beyond technical performance. In applied energy forecasting, particularly in domains like district heating where decisions affect consumers and operations in real time, trust in model predictions is critical. Practitioners and policymakers must have confidence that AI models are learning from representative, credible data. Visual data diagnostics provide a transparent and auditable method to assess data quality before model training, thereby increasing trust in both the modeling process and its outcomes. This is essential for the wider acceptance of AI tools in public infrastructure management.

In summary, visual data diagnostics and cleaning directly contribute to the reliability of AI-driven predictions. By ensuring that LSTM/GRU models train on representative, high-quality data, these models are enabled to deliver accurate forecasts of heat consumption. This is critical for district heating utilities as an accurate model helps in optimizing heat production (avoiding over- or under-production), integrating renewable heat sources efficiently, and ultimately providing stable indoor temperatures for consumers. The investment in data quality pays off in the form of more trustworthy and effective AI models.

## 7 Discussion

The presented approach demonstrates that visual diagnostics are a practical and powerful complement to automated data cleaning in the energy domain. A key advantage is the ability to leverage human expertise and intuition. Domain experts (such as DH network operators) can often spot when a pattern "doesn't look right" in a plot, drawing on contextual knowledge (e.g., knowing that heat demand should never be zero on a freezing day). By visualizing the data, these experts can inject domain knowledge into the data preparation process; something purely algorithmic methods might be missing. This collaborative human-in-the-loop strategy aligns with the broader trend in visual analytics to integrate human judgment in data quality management [7].

One might question how scalable visual diagnostics are, given that Danish DH systems involve thousands of meters, e.g. [2] presents a dataset containing over 30,000 meters. Manually inspecting each time series is infeasible. However, the techniques proposed can be scaled and combined with automated methods. For example, an anomaly detection algorithm could first flag meters or periods with potential issues, and then engineers use focused visualizations to examine those flagged subsets in detail. This guided approach greatly reduces the burden – instead of blindly searching through all data, analysts review specific plots where the algorithm suspects anomalies. Additionally, aggregated visualizations (like Fig. 2's heatmap of missing data across all meters) allow a system-level view that quickly highlights whether many devices have problems concurrently, pointing to systemic faults (e.g., a communication network downtime). In practice, a hierarchical approach can be used: start with high-level visuals (system-wide heatmaps, summary statistics plots), identify areas of concern, then drill down into individual metering plots as needed.

The insights from visual diagnostics can inform the design of automated data cleaning pipelines. For instance, if visual analysis frequently finds small negative blips that correspond to known meter firmware quirks, a rule can be added to the pipeline to auto-correct those (e.g., set any isolated negative reading to zero). In this case, after manually correcting a few such issues and confirming their nature, the fix could be generalized. Likewise, discovering that most missing data occur in short bursts might justify using linear interpolation as a default imputation in the pipeline for gaps under a certain length (supported by evidence in prior work [3]). Thus, visual diagnosis can lead to systematic improvements and more sophisticated data validation checks coded into the system.

District heating networks and consumption behavior evolve over time (e.g., insulation improvements, weather pattern changes, etc.), so models like LSTM/GRU are often retrained periodically or updated with new data. Continual monitoring of data quality is therefore essential. Visual diagnostics should not be a one-off exercise; utilities can establish periodic data quality reports with visual summaries. For example, each month an automated report could include plots of consumption vs. temperature for random samples of buildings, histograms of readings, and heatmaps of missing data. Any anomalies can be caught early and cleaned before retraining the AI models. This ensures that model performance does not degrade due to reduced data quality issues. It also helps maintain stakeholder trust: showing decision-makers that the data feeding AI models is consistently checked and clean can increase confidence in model-driven operational decisions.

While the focus is on Danish district heating data, the principles of visual data quality assessment are applicable to other energy systems and time-series sensor data. Electric load profiles, water usage data, and even industrial sensor readings share similar characteristics (periodicity, weather or production dependencies, occasional sensor errors). The visual techniques (line plots, heatmaps, scatter correlations, etc.) can be readily applied to those domains. It is anticipated that as Internet-of-Things deployments grow, combining human visual analysis with AI will be an important paradigm for ensuring data reliability. Limitations: It is important to acknowledge that visual diagnostics, while powerful, have limitations. Subtle errors or biases in data might not be immediately obvious visually. For example, a sensor calibration drift that causes a small systematic bias each day might not stand out in a plot, yet it could affect model predictions. Automated statistical tests and domain-specific knowledge are needed to catch such issues. Moreover, the effectiveness of visual methods depends on the skill and experience of the analyst. There is a risk of human bias – one might see a "pattern" in randomness or overlook an anomaly due to cognitive fatigue. To mitigate this, a combination of methods (visual + algorithmic) and peer review of findings can be used in practice. Another limitation is that visual fixes (like manually choosing to interpolate a gap) may not always be optimal; alternative imputation or smoothing techniques (e.g., forward-fill, spline interpolation, or using a model to impute) might sometimes perform better, and these choices need validation.

# 8 Conclusion

This paper presents a comprehensive approach to evaluating data quality for AI-driven heat consumption prediction in Danish district heating systems. By applying visual diagnostics through an interactive dashboard, the study demonstrates how time series plots, heatmaps, histograms, box plots, and correlation analyses can be used to identify missing values, outliers, and structural inconsistencies in smart meter data. A real-world case study shows how these visual tools enable data cleaning decisions that directly enhance the reliability of AI models such as LSTM and GRU.

The core contribution of this work lies in bridging raw sensor data and machine learning pipelines through a human-in-the-loop visual analytics process. The dashboard design, based on open-source Python tools and structured KPIs, provides a scalable and transparent mechanism for data quality assurance. This supports not only more accurate forecasts but also improved stakeholder trust in AI-generated predictions. In domains like district heating, where operational decisions affect consumer comfort and infrastructure efficiency, ensuring the integrity of input data is essential.

While the current study provides a conceptual framework and illustrative case application, future work should include empirical validation of the effects of data cleaning on AI model performance. In particular, comparing forecasting metrics such as RMSE and MAE between models trained on cleaned versus uncleaned datasets would provide quantitative evidence of the impact of visual diagnostics. Furthermore, while this study illustrates how anomalies and missing values can be identified and addressed through visual diagnostics, future work should formalize treatment strategies, such as imputation, removal, or automated handling and assess their impact on AI model performance.

Future research should also explore the integration of visual diagnostics with automated anomaly detection techniques to support real-time data quality monitoring. Extending the dashboard architecture to accommodate other types of utility data (such as electricity, water, or indoor climate sensors) will further demonstrate its generalizability. Finally, linking meter-level metadata such as building type or control strategy will enhance the explanatory power of the visual analytics and enable more targeted interventions.

## Acknowledgement

This paper is part of the projects titled "Automated Data and Machine Learning Pipeline for Cost-Effective Energy Demand Forecasting in Sector Coupling" (jr. Nr. RF-23-0039; Erhvervsfyrtårn Syd Fase 2), European Regional Development Fund and "Danish participation in IEA DHC Annex TS9 - Digitalization of District Heating and Cooling: Improving Efficiency and Performance Through Data Integration", funded by EUDP (project number: 95-41006-2410289).

## Declarations

**Conflict of Interest/Competing Interests:**
The authors declare no competing interests

**Ethics approval/Consent:**
Not applicable

**Data availability:**
Not applicable